# The Bayesian Structural EM Algorithm

**Nir Friedman**[*]
Computer Science Division, 387 Soda Hall
University of California, Berkeley, CA 94720
nir@cs.berkeley.edu

## Abstract

In recent years there has been a flurry of works on learning Bayesian networks from data. One of the hard problems in this area is how to effectively learn the structure of a belief network from *incomplete* data—that is, in the presence of missing values or hidden variables. In a recent paper, I introduced an algorithm called *Structural EM* that combines the standard Expectation Maximization (EM) algorithm, which optimizes parameters, with structure search for model selection. That algorithm learns networks based on penalized likelihood scores, which include the BIC/MDL score and various approximations to the Bayesian score. In this paper, I extend Structural EM to deal directly with Bayesian model selection. I prove the convergence of the resulting algorithm and show how to apply it for learning a large class of probabilistic models, including Bayesian networks and some variants thereof.

## 1 INTRODUCTION

*Belief networks* are a graphical representation for probability distributions. They are arguably the representation of choice for uncertainty in artificial intelligence and have been successfully applied in expert systems, diagnostic engines, and optimal decision making systems. Eliciting belief networks from experts can be a laborious and expensive process. Thus, in recent years there has been a growing interest in learning belief networks from data [9, 16, 17, 18]. Current methods are successful at learning both the structure and parameters from *complete* data—that is, when each data record describes the values of all variables in the network. Unfortunately, things are different when the data is *incomplete*. Until recently, learning methods were almost exclusively used for adjusting the parameters for a fixed network structure.

The inability to learn structure from incomplete data is considered as one of the main problems with current state of the art technology for several reasons. First, most real-life data contains *missing values* One of the cited advantages of belief networks (e.g., [16]) is that they allow for principled methods for reasoning with incomplete data. However, it is unreasonable at the same time to require complete data for training them. Second, learning a concise structure is crucial both for avoiding overfitting and for efficient inference in the learned model. By introducing *hidden* variables that do not appear explicitly in the model we can often learn simpler models.

In [12], I introduced a new method for searching over structures in the presence of incomplete data. The key idea of this method is to use our "best" estimate of the distribution to *complete* the data, and then use procedures that work efficiently for complete data on this completed data. This follows the basic intuition of the *Expectation Maximization* (EM) algorithm for learning parameters in a fixed parametric model [11]. Hence, I call this method *Structural EM*. (In [12], the name MS-EM was used.) Roughly speaking, Structural EM performs search in the joint space of (Structure × Parameters). At each step, it can either find better parameters for the current structure, or select a new structure. The former case is a standard "parametric" EM step, while the later is a "structural" EM step. In [12], I show that for penalized likelihood scoring functions, such as the BIC/MDL score [18], this procedure converges to a "local" maxima.

A drawback of the algorithm of [12] is that it applies only to scoring functions that approximate the Bayesian score. There are good indications, both theoretical and empirical, that the exact Bayesian score provides a better assessment of the generalization properties of a model given the data. Moreover, the Bayesian score provides a principled way of incorporating prior knowledge into the learning process.[1]

To compute the Bayesian score of a network, we need to integrate over all possible parameter assignments to the network. In general, when data is incomplete, this integral cannot be solved in closed form. Current attempts to learn from incomplete data using the Bayesian score use either stochastic simulation or Laplace's approximation to approximate this integral (see [7] and the references within). The former methods tend to be computationally expensive, and the latter methods can be imprecise. In particular, the Laplace approximation assumes that the likelihood function is unimodal, while there are cases where we know that this function has an exponential number of modes.

In this paper, I introduce a framework for learning probabilistic models using the Bayesian score under standard assumptions on the form of the prior distribution. As with Structural EM, this method is also based on the idea of completion of the data using our best guess so far. However, in this case the search is over the space of structures rather than the space of structures and parameters.

This paper is organized as follows. In Section 2, I describe a class of models, which I call *factored models*, that includes belief networks, multinets, decision trees, decision graphs, and many other probabilistic models. I review how to learn these from complete data and the problems posed by incomplete data. In Section 3, I describe the Bayesian Structural EM algorithm in a rather abstract settings and discuss its convergence properties. The algorithm, as presented in Section 3, cannot be directly implemented, and we need to approximate some quantities. In Section 4, I discuss how to adapt the algorithm for learning factored models. This results in an approximate approach that is different from the standard ones in the literature. It is still an open question whether it is more accurate. However, the derivation of this approximation is based on computational consideration of how to search in the space of network structures. Moreover, the framework I propose here suggests where possible improvements can be made. Finally, in Section 5, I describe experimental results that compare the performance of networks learned using the Bayesian Structural EM algorithm and networks learned using the BIC score.

---

[*]*Current address*: Institute of Computer Science, The Hebrew University, Givat Ram, Jerusalem 91904, Israel, nir@cs.huji.ac.il.

[1]It is worth noting that the Structural EM procedure, as presented in [12], is applicable to scores that include priors over parameters. Such scores incorporate, to some extent, the prior knowledge by learning MAP parameters instead of maximum likelihood ones.

## 2    PRELIMINARIES

In this section, I define a class of *factored models* that includes various variants of Bayesian networks, and briefly discuss how to learn them from complete and incomplete data, and the problems raised by the latter case.

### 2.1    FACTORED MODELS

We start with some notation. I use capital letters, such as $X, Y, Z$, for variable names and lowercase letters $x, y, z$ to denote specific values taken by those variables. Sets of variables are denoted by boldface capital letters $\mathbf{X}, \mathbf{Y}, \mathbf{Z}$, and assignments of values to the variables in these sets are denoted by boldface lowercase letters $\mathbf{x}, \mathbf{y}, \mathbf{z}$.

In learning from data we are interested in finding the best explanation for the data from a set of possible explanations. These explanations are specified by sets of hypotheses that we are willing to consider. We assume that we have a class of *models* $\mathcal{M}$ such that each model $M \in \mathcal{M}$ is parameterized by a vector $\Theta^M$ such that each (legal) choice of values $\Theta^M$ defines a probability distribution $\Pr(\cdot : M^h, \Theta^M)$ over possible data sets, where $M^h$ denotes the hypothesis that the underlying distribution is in the model $M$. (From now on I use $\Theta$ as a shorthand for $\Theta^M$ when the model $M$ is clear from the context.) I require that the intersection between models has zero measure, and from now on, we will treat $M^h$ and $M'^h$ as disjoint events.

We now examine conditions on $\mathcal{M}$ for which the algorithms described below are particularly useful.

The first assumption considers the form of models in $\mathcal{M}$. A *factored* model $M$ (for $\mathbf{U} = \{X_1, \ldots, X_n\}$) is a parametric family with parameters $\Theta^M = (\Theta_1^M, \ldots, \Theta_k^M)$ that defines a joint probability measure of the form:

$$\Pr(X_1, \ldots, X_n \mid M^h, \Theta^M) = \prod_i f_i^M(X_1, \ldots, X_n : \Theta_i^M),$$

where each $f_i^M$ is a *factor* whose value depends on some (or all) of the variables $X_1, \ldots, X_n$. A factored model is *separable* if the space of legal choices of parameters is the cross product of the legal choices of parameters $\Theta_i^M$ for each $f_i^M$. In other words, if legal parameterization of different factors can be combined without restrictions.

**Assumption 1.** All the models $\mathcal{M}$ are separable factored models.

This assumption by itself is not too strong, since any probability model can be represented by a single factor. Here are some examples of non-trivially factored models that are also separable.

**Example 2.1:** A *belief network* [22] is an annotated directed acyclic graph that encodes a joint probability distribution over $\mathbf{U}$. Formally, a belief network for $\mathbf{U}$ is a tuple $B = \langle G, \mathcal{L}, \Theta \rangle$. The first component, namely $G$, is a directed acyclic graph whose vertices correspond to the random variables $X_1, \ldots, X_n$ that encodes the following set of conditional independence assumptions: each variable $X_i$ is independent of its non-descendants given its parents in $G$. The second component of the tuple, namely $\mathcal{L}$, is a set of *local models* $L_1, \ldots, L_n$. Each local model $L_i$ maps possible values $\mathbf{pa}(X_i)$ of $\mathbf{Pa}(X_i)$, the set of parents of $X_i$, to a probability measure over $X_i$. The local models are parameterized by parameters $\Theta_i$. A belief network $B$ defines a unique joint probability distribution over $\mathbf{U}$ given by:

$$P_B(X_1, \ldots, X_n) = \prod_{i=1}^{n} L_i(X_i, \mathbf{Pa}(X_i) : \Theta_i)$$

It is straightforward to see that a belief network is a factored model. Moreover, it is separable: since any combination of locally legal parameters defines a probability measure. ∎

**Example 2.2:** As a more specific example, consider belief networks over variables that have a finite set of values. A standard representation of the local models in such networks is by a table. For each assignment of values to $\mathbf{Pa}(X_i)$, the table contains a conditional distribution over $X_i$. In such networks, we can further decompose each of the local models into a product of *multinomial* factors: $\prod_{\mathbf{pa}(X_i)} L_{i,\mathbf{pa}(X_i)}(X_i, \mathbf{Pa}(X_i) : \Theta_{i,\mathbf{pa}(X_i)})$, where $\Theta_{i,\mathbf{pa}(X_i)}$ is a vector that contains parameters $\theta_{x_i,\mathbf{pa}(X_i)}$ for each value $x_i$ of $X_i$, and $L_{i,\mathbf{pa}(X_i)}(X_i, \mathbf{Pa}(X_i) : \Theta_{i,\mathbf{pa}(X_i)})$ is $\theta_{x_i,\mathbf{pa}(X_i)}$ if $\mathbf{Pa}(X_i) = \mathbf{pa}(X_i)$ and $X_i = x_i$, and 1 otherwise. In this case, we can write the joint probability distribution $P_B(X_1, \ldots, X_n \mid M, \Theta^M)$ as

$$\prod_{i=1}^{n} \prod_{\mathbf{pa}(X_i)} L_{i,\mathbf{pa}(X_i)}(X_i, \mathbf{Pa}(X_i) : \Theta_{i,\mathbf{pa}(X_i)}).$$

Again, it is easy to verify that such a model is separable: each combination of legal choices of $\Theta_{i,\mathbf{pa}(X_i)}$ results in a probability distribution. ∎

Other examples of separable factored models include multinets [14], mixture models [6], decision trees [5], decision graphs, and the combination of the latter two representations with belief networks [4, 13, 8]. An example of a class of models that are factored in a non-trivial sense but are not separable are non-chordal Markov networks [22]. The probability distribution defined by such networks has a product form. However, a change in the parameters for one factor requires changing the global normalizing constant of the model. Thus, not every combination of parameters results in a legal probability distribution.

Our next assumption involves the choice of factors in the factored models. I require that each factor is from the *exponential family* [10]: A factor is *exponential* if it can be specified in the form

$$f(\mathbf{X} : \Theta) = e^{t(\Theta) \cdot s(\mathbf{X})}$$

where $t(\Theta)$ and $s(\mathbf{X})$ are vector valued functions of the same dimension, and $\cdot$ is the inner product. [2]

**Example 2.3:** It is easy to verify that the multinomial factors from Example 2.2 are exponential. We can rewrite

---

[2] Standard definitions of the exponential family often include an additional normalizing term and represent the distribution as $a(\Theta)e^{t(\Theta) \cdot s(\mathbf{X})}$. However, this term can be easily accounted for by adding an additional dimension to $t(\cdot)$ and $s(\mathbf{X})$.

$L_{i,\mathbf{pa}(X_i)}(X_i, \mathbf{Pa}(X_i) : \Theta_{i,\mathbf{pa}(X_i)})$ in the exponential form by setting

$$t(\Theta_{i,\mathbf{pa}(X_i)}) = \langle \log \theta_{v_1,\mathbf{pa}(X_i)}, \ldots, \log \theta_{v_l,\mathbf{pa}(X_i)} \rangle$$
$$s(\mathbf{x}) = \langle 1_{v_1,\mathbf{pa}(X_i)}(\mathbf{x}), \ldots, 1_{v_l,\mathbf{pa}(X_i)}(\mathbf{x}) \rangle$$

where $v_1, \ldots, v_l$ are the possible values of $X_i$, and $1_{\mathbf{y}}(\mathbf{x})$ if the values of $\mathbf{Y} \subseteq \mathbf{X}$ in $\mathbf{y}$ match the values assigned to them by $\mathbf{x}$, and 0 otherwise. ∎

Other examples of exponential factors include univariate and multivariate Gaussians, and many other standard distributions (see, for example, [10]).

**Assumption 2.** All the models in $\mathcal{M}$ contain only exponential factors.

## 2.2 BAYESIAN LEARNING

Assume that we have an input dataset $D$ with some number of examples. We want to predict other events that were generated from the same distribution as $D$. To define the Bayesian learning problem, we assume that learner has a prior distribution over models $\Pr(M^h)$, and over the parameters for each model, $\Pr(\Theta^M \mid M^h)$. Bayesian learning attempts to make predictions by conditioning the prior on the observed data. Thus, the prediction of the probability of an event $X$, after seeing the training data, can be written as:

$$\Pr(X \mid D) = \sum_M \Pr(X \mid M^h, D) \Pr(M^h \mid D)$$
$$= \sum_M \Pr(X \mid M^h, D) \frac{\Pr(D \mid M^h) \Pr(M^h)}{\Pr(D)} \quad (1)$$

where

$$\Pr(D \mid M^h) = \int \Pr(D \mid M^h, \Theta) \Pr(\Theta \mid M^h) d\Theta, \quad (2)$$

and

$$\Pr(X \mid M^h, D) = \int \Pr(X \mid M^h, \Theta) \Pr(\Theta \mid M^h, D) d\Theta. \quad (3)$$

Usually, we cannot afford to sum over all possible models. Thus, we approximate (1) by using only the *maximum a posteriori* (MAP) model, or using a sum over several of the models with highest posterior probabilities. This is justified when the data is sufficient to distinguish among models, since then we would expect the posterior distribution to put most of the weight on a few models.

## 2.3 LEARNING FROM COMPLETE DATA

When the data is complete, that is, each example in $D$ assigns value to all the variables in $\mathbf{U}$, then learning can exploit the factored structure of models. To do so, we need to make assumptions about the prior distributions over the parameters in each model. We assume that *a priori*, the parameters for each factor are independent of the parameters of all other factors and depend only on the form of the factor. These assumptions are called *parameter independence* and *parameter modularity* by Heckerman et al. [17].

**Assumption 3.** For each model $M \in \mathcal{M}$ with $k$ factors the prior distribution over parameters has the form

$$\Pr(\Theta_1^M, \ldots, \Theta_k^M \mid M^h) = \prod_i \Pr(\Theta_i^M \mid M^h).$$

**Assumption 4.** If $f_i^M = f_j^{M'}$ for some $M, M' \in \mathcal{M}$, then $\Pr(\Theta_i^M \mid M^h) = \Pr(\Theta_j^{M'} \mid M'^h)$.

Given Assumptions 3 and 4, we can denote the prior over parameters of a factor $f_i$ as $\Pr(\Theta_i)$.

In practice, it also useful to require that the prior for each factor is a *conjugate* prior. For example, Dirichlet priors are conjugate priors for multinomial factors. For many types of exponential distributions, the conjugate priors lead to a closed-form solution for the posterior beliefs, and for the probability of the data.

An important property of learning given these four assumptions is that the probability of complete data given the model also has a factored form that mirrors the factorization of the model.

**Proposition 2.4:** *Given Assumptions 1–4 and a data set $D = \{\mathbf{u}^1, \ldots, \mathbf{u}^N\}$ of complete assignments to $\mathbf{U}$, the score of a model $M$ that consists of $k$ factors $f_1, \ldots, f_k$, is*

$$\Pr(D \mid M^h) = \prod_{i=1}^k F_i \left( \sum_{j=1}^N s_i(\mathbf{u}^j) \right),$$

*where*

$$F_i(S) = \int e^{t_i(\Theta_i) \cdot S} \Pr(\Theta_i) d\Theta_i,$$

*and $t_i(\cdot)$, and $s_i(\cdot)$ are the the exponential representation of $f_i$.*

It important to stress that terms in the score of Proposition 2.4 depend only on accumulated sufficient statistics in the data. Thus, to evaluate the score of a model, we can use a summary of the data in the form of accumulated sufficient statistics.

**Example 2.5** We now complete the description of the learning problem of multinomial belief networks. Following [9, 17] we use *Dirichlet priors*. A Dirichlet prior for a multinomial distribution of a variable $X$ is specified by a set of *hyperparameters* $\{N'_{v_1}, \ldots, N'_{v_l}\}$ where $v_1, \ldots, v_l$ are the values of $X$. We say that

$$\Pr(\Theta) \sim \text{Dirichlet}(\{N'_{v_1}, \ldots, N'_{v_l}\}) \text{ if } \Pr(\Theta) \propto \prod_{v_i} \theta_{v_i}^{N'_{v_i}-1}.$$

For a Dirichlet prior with parameters $N'_{v_1}, \ldots, N'_{v_k}$ the probability of the values of $X$ with sufficient statistics $S = \langle N_{v_1}, \ldots, N_{v_k} \rangle$ is given by

$$F(S) = \frac{\Gamma(\sum_i N'_{v_i})}{\Gamma(\sum_i (N'_{v_i} + N_{v_i}))} \prod_i \frac{\Gamma(N'_{v_i} + N_{v_i})}{\Gamma(N'_{v_i})}, \quad (4)$$

where $\Gamma(x) = \int_0^\infty t^{x-1} e^{-t} dt$ is the *Gamma* function. For more details on Dirichlet priors, see [10].

Thus, to learn multinomial Bayesian networks with Dirichlet priors, we only need to keep counts of the form $N_{x,\mathbf{pa}(X_x)}$ for families we intend to evaluate. The score of the network is a product of terms of the form of (4), one for each multinomial factor in the model; see [9, 17]. A particular score of this form is the BDe score of [17], which we use in the experiments below. ∎

Learning factored models from data is done by searching over the space of models for a model (or models) that maximizes the score. The above proposition shows that if we change a factored model *locally*, that is by replacing a few of the factors, then the score of the new model differs from the score of the old model by only a few terms. Moreover, by caching accumulated sufficient statistics for various factors, we can easily evaluate various combinations of different factors.

**Example 2.6:** Consider the following examples of search procedures that exploit these properties. The first is the search used by most current procedures for learning belief networks from complete data. This search procedure considers all arc additions, removals and reversals. Each of these operations changes only the factors that are involved in the conditional probabilities distributions of one or two variables. Thus, to execute a hill climbing search, we have to consider approximately $O(n^2)$ neighbors for at each point in the search. However, the change in the score due to one local modification remains the same if we modified another, unrelated, part of the network. Thus, at each step, the search procedure needs only to evaluate the $O(n)$ modifications that involve further changes to the parts of the model that were changed in the previous iteration.

Another example of a search procedure that exploits the same factorization properties is the standard "divide and conquer" approach for learning decision trees, see for example [5]. A decision tree is a factored model where each factor corresponds to a leaf of the tree. If we replace a leaf by subtree, or replace a subtree by a leaf, all of the other factors in the model remain unchanged. This formal property justifies independent search for the structure of each subtree once we decide the root of the tree. ∎

## 2.4    LEARNING FROM INCOMPLETE DATA

Learning factored models from incomplete data is harder than learning from complete data. This is mainly due to the fact that the posterior over parameters is no longer a product of independent terms. For the same reason, the probability of the data is no longer a product of terms.

Since the posterior distribution over the parameters of a model is no longer a product of independent posteriors, we usually cannot represent it in closed form. This implies that we cannot make exact predictions given a model using the integral of (3). Instead we can attempt to approximate this integral. The simplest approximation is by using MAP parameters. Roughly speaking, if we believe that the posterior over parameters is sharply peaked, than the integral in (3) is dominated by the predication in a small region around the posterior's peak. Thus, we approximate

$$\Pr(X \mid M^h, D) \approx \Pr(X \mid M^h, \hat{\Theta}) \qquad (5)$$

where $\hat{\Theta}$ is the vector of parameters that maximizes $\Pr(\Theta \mid M^h, D) \propto \Pr(D \mid \Theta, M^h) \Pr(\Theta \mid M^h)$. We can find an approximation to these parameters using either gradient ascent methods [3] or using EM [11, 19].

Since the probability of the data given a model no longer decomposes, we need to directly estimate the integral of (2). We can do so either using stochastic simulation, which is extremely expensive in terms of computation, or using large-sample approximations that are based on Laplace's approximation. The latter approximation assumes that posterior over parameters is peaked, and use a Gaussian fit in the neighborhood of the MAP parameters to estimate the integral. We refer the reader to [7, 15] for a discussion of approximations based on this technique.

The use of these approximations requires us to find the MAP parameters for each model we want to consider before we can score it. Thus, a search of model space requires an expensive evaluation of each candidate. When we are searching in a large space of possible models, this type of search becomes infeasible—the procedure has to invest a large amount of computation before making a single change in the model. Thus, although there have been thorough investigations of the properties of various approximations to the Bayesian score, there have been few empirical reports of experiments with learning structure, except in domains where the search is restricted to a small number of candidates (e.g., [6]).

## 3    THE STRUCTURAL EM ALGORITHM

In this section, I present the Bayesian Structural EM algorithm for structure selection. This algorithm attempts to directly optimize the Bayesian score rather than an asymptotic approximation. The presentation is in a somewhat more general settings than factored models. In the next section, we will see how to specialize it to factored models.

Assume that we have an input dataset $D$ with some number of examples. For the rest of this section, assume that the dataset is fixed, and denote each value, either supplied or missing, in the data by a random variable. For example, if we are dealing with a standard learning problem where the training data consists of $N$ i.i.d. instances, each of which is, a possibly partial assignment to $k$ variables, then we have $kN$ random variables that describe the training data. I denote by $\mathbf{O}$ the set of observable variables; that is, the set of variables whose values are determined by the training data. Similarly, I denote by $\mathbf{H}$ be the set of hidden (or unobserved) variables, that is, the variables that are not observed.

As before, we assume that we have a class of models $\mathcal{M}$ such that each model $M \in \mathcal{M}$ is parameterized by a vector $\Theta^M$ such that each (legal) choice of values $\Theta^M$ defines a probability distribution $\Pr(\cdot : M, \Theta^M)$ over $\mathbf{V}$. We also assume that we have a prior over models and parameter assignments in each model. For the sake of clarity, the following discussion assumes that all variables take values from a a finite set. However, the results in this section easily apply to continuous variables, if we make standard continuity and smoothness restrictions on the likelihood functions of models in $\mathcal{M}$.

To find a MAP model it suffices to maximize $\Pr(D \mid M^h) \Pr(M^h)$, since the normalizing term $Pr(D)$ is the same for all the models we compare. As we have seen in the previous section, if $D$ contains missing values, then we usually cannot evaluate $\Pr(D \mid M^h)$ efficiently. For the following discussion we assume that we can compute or estimate the complete data likelihood, $\Pr(\mathbf{H}, \mathbf{O} \mid M^h)$. As we have seen in the previous section, this assumption is true for the class of factored models satisfying Assumptions 1–4. We will also assume that given a particular model, we can perform the predictive inference of (3) efficiently. As we have seen, although this is not true for factored models, we can efficiently compute approximations for these predictions (e.g., using the MAP approximation).

We now have the tools to describe the general outline of the Bayesian Structural EM algorithm.

Procedure Bayesian-SEM($M_0$, **o**):
  Loop for $n = 0, 1, \ldots$ until convergence
    Compute the posterior $\Pr(\Theta^{M_n} \mid M_n^h, \mathbf{o})$.
    **E-step:** For each $M$, compute
$$Q(M : M_n) = E[\log \Pr(\mathbf{H}, \mathbf{o}, M^h) \mid M_n^h, \mathbf{o}]$$
$$= \sum_{\mathbf{h}} \Pr(\mathbf{h} \mid \mathbf{o}, M_n^h) \log \Pr(\mathbf{h}, \mathbf{o}, M^h)$$
    **M-step** Choose $M_{n+1}$ that maximizes $Q(M : M_n)$
    if $Q(M_n : M_n) = Q(M_{n+1} : M_n)$ then
      return $M_n$

The main idea of this procedure is that at each iteration it attempts to maximize the expected score of models instead of their actual score. There are two immediate questions to ask. Why is this easier? and, what does it buy us? The answer to the first question depends on the class of models we are using. As we shall see below, we can efficiently evaluate the expected score of factored models.

We now address the second question. The following theorem shows that procedure makes "progress" in each iteration.

**Theorem 3.1:** *Let* $M_0, M_1, \ldots$ *be the sequence of models examined by the Bayesian SEM procedure. Then,*

$$\log \Pr(\mathbf{o}, M_{n+1}^h) - \log \Pr(\mathbf{o}, M_n^h)$$
$$\geq Q(M_{n+1} : M_n) - Q(M_n : M_n)$$

**Proof:**

$$\log \frac{\Pr(\mathbf{o}, M_{n+1}^h)}{\Pr(\mathbf{o}, M_n^h)}$$

$$= \log \sum_{\mathbf{h}} \frac{\Pr(\mathbf{h}, \mathbf{o}, M_{n+1}^h)}{\Pr(\mathbf{o}, M_n^h)} \cdot \frac{\Pr(\mathbf{h}|\mathbf{o}, M_n^h)}{\Pr(\mathbf{h}|\mathbf{o}, M_n^h)}$$

$$= \log \sum_{\mathbf{h}} \Pr(\mathbf{h} \mid \mathbf{o}, M_n^h) \frac{\Pr(\mathbf{h}, \mathbf{o}, M_{n+1}^h)}{\Pr(\mathbf{h}, \mathbf{o}, M_n^h)} \quad (6)$$

$$\geq \sum_{\mathbf{h}} \Pr(\mathbf{h} \mid \mathbf{o}, M_n^h) \log \frac{\Pr(\mathbf{h}, \mathbf{o}, M_{n+1}^h)}{\Pr(\mathbf{h}, \mathbf{o}, M_n^h)} \quad (7)$$

$$= E[\log \frac{\Pr(\mathbf{H}, \mathbf{o}, M_{n+1}^h)}{\Pr(\mathbf{H}, \mathbf{o}, M_n^h)} \mid M_n^h, \mathbf{o}]$$

$$= Q(M_{n+1} : M_n) - Q(M_n : M_n)$$

where all the transformations are by algebraic manipulations, and the inequality between (6) and (7) is a consequence of Jensen's inequality.[3] ∎

This theorem implies that if $Q(M_n : M_{n+1}) > Q(M_n : M_n)$ then $\Pr(\mathbf{o}, M_{n+1}^h) > \Pr(\mathbf{o}, M_n^h)$. Thus, if we choose a model that maximizes the expected score at each iteration, then we are *provably* making a better choice, in terms of the marginal score of the network. It is important to note that this theorem also implies that we can use a weaker version of the **M-step**:

**M\*-step** Choose $M_{n+1}$ such that
$$Q(M_{n+1} : M_n) > Q(M_n : M_n)$$

This is analogous to the *Generalized EM* algorithm. Using this variant, we do not need to evaluate the expected score of all possible models in the **E-Step**. In fact, as we shall see below, in practice we only evaluate the expected score of a small subset of the models.

Theorem 3.1 implies that the procedure converges when there is no further improvement in the objective score. As an immediate consequence, we can show that the procedure reaches such a point under fairly general conditions.

**Theorem 3.2:** *Let* $M_0, M_1, \ldots$ *be the sequence of models examined by the Bayesian SEM procedure. If the number of models in $\mathcal{M}$ is finite, or if there is a constant $c$ such that* $\Pr(D \mid M^h, \Theta^M) < c$ *for all models $M$ and parameters $\Theta^M$, then the limit* $\lim_{n \to \infty} \Pr(\mathbf{o}, M_n^h)$ *exists.*

Unfortunately, there is not much we can say about the convergence points. Recall that for the standard EM algorithm, convergence points are stationary points of the objective function. There is no corresponding notion in the discrete space of models we are searching over. In fact, the most problematic aspect of this algorithm is that it might converge to a sub-optimal model. This can happen if the model generates a distribution that makes other models appear worse when we examine the expected score. Intuitively, we would expect this phenomena to become more common as the ratio of missing information is higher. In practice we might want to run the algorithm from several starting points to get a better estimate of the MAP model.

## 4 BAYESIAN STRUCTURAL EM FOR FACTORED MODELS

We now consider how to apply the Bayesian Structural EM algorithm for factored models. There are several issues that we need to address in order to translate the abstract algorithm into a concrete procedure.

Recall that each iteration of the algorithm requires the evaluation of the expected score $Q(M : M_n)$ for each model we examine. Since the term inside the expected score involves assignments to **H**, we can evaluate $\Pr(\mathbf{h}, \mathbf{o} \mid M^h)$ as though we had complete data. Using Proposition 2.4 and linearity of expectation we get the following property.

**Proposition 4.1:** *Let* $D = \{\mathbf{x}^1, \ldots, \mathbf{x}^N\}$ *be a training set that consist of incomplete assignments to $\mathbf{U}$. Given Assumptions 1–4, if $M$ consists of $k$ factors, $f_1, \ldots, f_k$, then*

$$E[\log \Pr(\mathbf{H}, \mathbf{o} \mid M^h)] = \sum_{i=1}^{k} E[\log F_i(S_i)],$$

*where $S_i = \sum_{j=1}^{N} s_i(\mathbf{U}^j)$ is a random variable that represents the accumulated sufficient statistics for the factor $f_i$ in possible completions of the data.*

An immediate consequence of this proposition is that the expected score has the same decomposability properties as the score with complete data—local changes to the model result in changes in only a few terms in the score. Thus, we can use complete data search procedures that exploit this property, such as the ones discussed in Example 2.6.

Next, we address the evaluation of terms of the form $E[\log F_i(S_i)]$. Here we have few choices. The simplest approximation has the form

$$E[\log F_i(S_i)] \approx \log F_i(E[S_i]) \quad (8)$$

This approximation is exact if $\log F_i(\cdot)$ is *linear* in its arguments. Unfortunately, this is not the case for members of the exponential family. Nonetheless, in some cases this approximation can be reasonably accurate. In other cases, we can correct for the non-linearity of $\log F_i(\cdot)$. In the next section, I expand on these issues and outline possible approximations of $E[\log F_i(S_i)]$. All of these approximations use $E[S_i]$ and some of them also use the covariance matrix of the vector $S$.

Computing these expectations (and variances) raises the next issue: How to compute the probability over assignments to **H**? According to the Bayesian-SEM procedure, we need to use $\Pr(\mathbf{H} \mid \mathbf{o}, M_n^h)$. However, as we discussed above, when we have incomplete data, we usually cannot evaluate this posterior efficiently. For now, we address this problem using the MAP approximation of (5). Thus, when we want to compute expectation based on $M_n$, we attempt to learn MAP parameters for $M_n$ and use these. This approximation is fairly standard and can be done quite efficiently. The computation of the MAP parameters can be done using either EM (as done in the experiments described below), gradient ascent or extensions of these methods. Moreover, once we fix the MAP parameters, we can use standard inference procedure using the model $(M_n, \hat{\Theta})$.[4]

---

[3]The same proof carries over to the case of continuous variables. We simply replace the summation over **h** with an integration. To apply Jensen's inequality we have to make some mild assumptions on the density function defined by models in $\mathcal{M}$.

[4]We must remember, however, that this approximation is imprecise, since it ignores most of the information of the posterior. A possible way of improving this approximation is by considering a better approximation of the posterior, such as *ensemble methods* [20].

When we use the MAP approximation, we get a procedure with the following structure:

Procedure Factored-Bayesian-SEM($M_0$, o):
  Loop for $n = 0, 1, \dots$ until convergence
    Compute the MAP parameters $\hat{\Theta}^{M_n}$ for $M_n$ given o.
    Perform search over models, evaluating each model by
      $Score(M : M_n) = \sum_i E[\log F_i^M(S_i^M) \mid o, M_n^h, \hat{\Theta}_n^M]$
    Let $M_{n+1}$ be the model with the highest score among
      these encountered during the search.
    if $Score(M_n : M_n) = Score(M_{n+1} : M_n)$ then
      return $M_n$

To completely specify this procedure we have to decide on the search method over structures. This depends on the class of models we are interested in. In some classes of models, such as the class of Chow trees, there are algorithms that construct the best scoring model. (See [21] for a nice use of this idea within an approach that is similar to Structural EM.) In other cases, we must resort to a heuristic search procedure, such as the ones discussed above. In general, any search procedure that exploits the decomposition properties of factored models in complete data can be used within the Factored-Bayesian-SEM algorithm.

Finally, as mentioned above, we need to estimate moments (e.g., mean and variance) of the distribution of $S_i$ in order to evaluate the score of a factor $f_i$. If many models share similar factors, we can cache the results of these computations. As a consequence, the evaluation of many models does not require additional inference. In some cases, we can schedule computation in advance, if we know which factors we will be examined during the search. A simple example of this idea is, again, the algorithm for learning Chow trees. In this case, we know in advance that we need to evaluate all factors that involve pairwise interactions between variables. Thus, we can compute the necessary information in one pass over the training data. (Again, see [21] for a nice use of this idea.) In addition the caching strategy can use the fact that for many classes of exponential families, such as multinomials and Gaussians, we can *marginalize* the sufficient statistics for one factor from these of another factor.

The upshot of this discussion is that we can use efficient search techniques inside the Bayesian Structural EM loop. These search algorithms can evaluate many candidates, since most candidates they explore share many factors. Thus, each new candidate might require evaluation of the expected score of only a few factors. In many cases, examining a new model requires no new factors to be evaluated.

### 4.1  COMPUTING $E[\log F(S)]$

We now examine how to approximate the value of $E[\log F(S)]$. For the purpose of this discussion assume that the factor in question is fixed and we omit the denote by $t(\cdot)$, $s(\cdot)$ and $F(\cdot)$ the associated functions.

We start our analysis by examining the distribution over the accumulated sufficient statistics $\bar{S}$. Recall that $S$ is a sum of the form $\sum_j s(\mathbf{U}^j)$, where $\mathbf{U}^j$ denotes the completion of the $j$'th instance under possible completions of the data. Since the joint distribution defined by any model over $\mathbf{H}$ is a product of independent distributions, one for each instance in the data, we have that the variables $s(\mathbf{U}^j)$ are independent. Using the central limit theorem we have that the distribution of $S$ can be approximated by a Gaussian distribution with mean $E[S] = \sum_j E[s(\mathbf{U}^j)]$, and covariance matrix $\Sigma[S] = \sum_j \Sigma[s(\mathbf{U}^j)]$. Both of these can be accumu-lated by performing some computation on each instance in the training data. Usually, we can compute the covariance matrix based on the same computations we use in order to compute the expected sufficient statistics

This observation implies that the distribution of $S$ becomes sharply peaked as the expected number of "effective" samples in the data grows. The "effective" samples are samples whose probability is sensitive to changes in the parameters of the factor. Formally, these are samples for which $s(\mathbf{U}^j)$ is not zero. For example, when learning multinomial Bayesian networks, the effective samples for the factor $L_{s,\mathbf{pa}(X_i)}$ are these where $\mathbf{Pa}(X_i) = \mathbf{pa}(X_i)$ (or can be assigned that value in some completions of the data).

As mentioned above, the simplest approximation of $E[\log F(S)]$ is using (8). This approximation is precise if $\log F(S)$ is linear in $S$. It can be fairly accurate if $\log F(S)$ can be approximated by linear function in the vicinity of $E[S]$. Since most of the density is assigned to values of $S$ in this region, this results in a good approximation. Formally, using Taylor expansion, to get that:

$$\log F(S) = \log F(E[S]) + (S - E[S])\nabla(\log F)(E[S]) + \tfrac{1}{2}(S - E[S])^T\nabla^2(\log F)(S^*)(S - E[S])$$

where $S^*$ is a point along the line from $E[S]$ to $S$. When we take expectation over the right hand side, the second term cancels out. Thus, the difference between $E[\log F(S)]$ and $\log F(E[S])$, is the integration of the quadratic term in the Taylor expansion. If we can show that the *norm* of the Hessian $\nabla^2(\log F)$ is bounded in the region of high density around $E[S]$, then we can bound the error.

My conjecture is that for factors from the *regular* exponential family, the norm of the Hessian asymptotes to 0, as the expected number of effective samples for $S$ grows. This is easily verified for multinomial factors. In this case, using simple approximation to the derivatives of $\log \Gamma(\cdot)$, we get that the elements of the Hessian are roughly of the form $\frac{1}{N_{v_i}} - \frac{1}{\sum_i N_{v_i}}$. Thus, as the size of the expected counts grows, the Hessian matrix vanishes. This implies for multinomial factors, in cases where the expected counts are far from 0, we can safely use the linear approximation of (8). I hope to provide a more definitive characterization of the conditions under which this approximation is close in the full version of this paper.

In cases where the linear approximation to $\log F(\cdot)$ does not suffice, we can get a better approximation by using the Gaussian approximation to the distribution over the values of $S$. Thus, we can approximate $E[\log F(S)]$ by an integral over a Gaussian

$$E[\log F(S)] \approx \int \log F(S)\varphi(S : E[S], \Sigma[S])dS, \quad (9)$$

where $\varphi(\mathbf{X} : \mu, \Sigma)$ is the multivariate Gaussian with mean $\mu$ and covariance matrix $\Sigma$. Note that the central limit theorem implies that the normal approximation is fairly good even for relatively small number of instances.

There are several methods for evaluating the right-hand side of (9). If the dimension of $S$ is small, we can use numerical integration techniques to directly evaluate the integral. If the dimension of $S$ is large, we can use Laplace's approximation. Here we have good reasons to believe that, if $\log F(\cdot)$ is well-behaved, then the integration is over a unimodal function, and therefore Laplace's approximation would work well. To perform Laplace's approximation in this case, we need to find the maximum point of $G(S) = \log F(S)\varphi(S : E[S], \Sigma[S])$ and then evaluate the Hessian of $\log G(S)$ at that point. The first step can be done by standard optimization methods (e.g., gradient ascent), and the second is a straight forward application of Laplace's approximation. Due to lack of space, I do not go in to details.

| Method | alarm | | | | insurance | | | |
|---|---|---|---|---|---|---|---|---|
| | 500 | 1000 | 2000 | 4000 | 500 | 1000 | 2000 | 4000 |
| **.10** | | | | | | | | |
| BDe (S) | 1.046+-.1210 | 0.504+-.0596 | 0.315+-.0423 | 0.214+-.0238 | 1.600+-.1042 | 1.075+-.0652 | 0.750+-.1205 | 0.449+-.0423 |
| BDe (I) | 1.151+-.0435 | 0.603+-.0888 | 0.337+-.0754 | 0.247+-.0147 | 1.855+-.1173 | 1.336+-.0727 | 0.889+-.1521 | 0.516+-.0839 |
| BDe (La) | 1.251+-.0933 | 0.841+-.1309 | 0.372+-.0541 | 0.269+-.0312 | 2.099+-.1485 | 1.634+-.1279 | 0.939+-.0875 | 0.825+-.1806 |
| BDe (Li) | 1.135+-.0741 | 0.566+-.0628 | 0.283+-.0264 | 0.257+-.0104 | 1.893+-.1442 | 1.296+-.1105 | 0.842+-.1531 | 0.543+-.0826 |
| BIC | 2.784+-.1779 | 1.257+-.1758 | 0.628+-.0857 | 0.594+-.0397 | 2.965+-.2642 | 1.850+-.1543 | 1.446+-.1449 | 0.950+-.0961 |
| **.20** | | | | | | | | |
| BDe (S) | 1.532+-.2158 | 0.724+-.0796 | 0.439+-.0894 | 0.259+-.0056 | 2.135+-.2018 | 1.623+-.0845 | 1.103+-.1435 | 0.668+-.0810 |
| BDe (I) | 1.581+-.2534 | 0.995+-.0655 | 0.634+-.0820 | 0.282+-.0848 | 2.328+-.1017 | 1.933+-.1418 | 1.423+-.0545 | 0.721+-.0749 |
| BDe (La) | 1.985+-.2114 | 0.984+-.1510 | 0.645+-.0364 | 0.470+-.1002 | 2.879+-.2236 | 2.069+-.3054 | 1.599+-.2313 | 0.819+-.0785 |
| BDe (Li) | 1.476+-.2226 | 1.056+-.0908 | 0.614+-.0630 | 0.228+-.0348 | 2.391+-.3829 | 1.791+-.1933 | 1.323+-.2199 | 0.796+-.1157 |
| BIC | 3.171+-.4608 | 1.870+-.1891 | 0.900+-.1863 | 0.564+-.0298 | 3.453+-.2542 | 2.614+-.1835 | 1.975+-.0730 | 1.490+-.1148 |
| **.30** | | | | | | | | |
| BDe (S) | 2.173+-.1349 | 1.239+-.1555 | 0.754+-.1098 | 0.455+-.1770 | 2.974+-.3019 | 2.211+-.0769 | 1.859+-.2894 | 1.196+-.2880 |
| BDe (I) | 2.683+-.3791 | 1.482+-.2893 | 0.832+-.0636 | 0.411+-.1049 | 3.515+-.3060 | 2.226+-.1221 | 2.046+-.1391 | 1.379+-.1801 |
| BDe (La) | 3.416+-.3835 | 1.576+-.2279 | 1.008+-.1685 | 0.675+-.0611 | 3.515+-.1865 | 2.781+-.3146 | 1.923+-.1734 | 1.511+-.1739 |
| BDe (Li) | 2.866+-.3641 | 1.685+-.1504 | 1.021+-.1724 | 0.579+-.1531 | 3.473+-.3690 | 2.475+-.1619 | 2.039+-.1147 | 1.634+-.2823 |
| BIC | 3.942+-.3839 | 3.131+-.1883 | 1.866+-.1700 | 4.126+-.0950 | 4.126+-.3303 | 3.320+-.3162 | 2.156+-.1297 | 1.874+-.1209 |
| **.40** | | | | | | | | |
| BDe (S) | 3.852+-.5568 | 2.192+-.3096 | 1.255+-.1653 | 1.794+-1.8763 | 4.342+-.5313 | 3.181+-.3114 | 2.024+-.1074 | 1.945+-.1730 |
| BDe (I) | 4.430+-.1813 | 2.564+-.4480 | 1.690+-.2122 | 1.824+-1.8615 | 4.320+-.5381 | 3.289+-.4039 | 2.238+-.1617 | 2.130+-.1716 |
| BDe (La) | 4.429+-.2635 | 3.038+-.3359 | 1.887+-.2115 | 1.006+-.1781 | 4.416+-.5386 | 3.246+-.4745 | 2.778+-.3226 | 2.017+-.1206 |
| BDe (Li) | 4.550+-.2485 | 3.061+-.3884 | 1.553+-.2431 | 0.740+-.1217 | 4.946+-.4052 | 3.584+-.4422 | 2.345+-.1130 | 2.025+-.0769 |
| BIC | 5.645+-.6852 | 3.821+-.0919 | 2.883+-.4775 | 1.549+-.2079 | 6.054+-.1423 | 3.714+-.2343 | 2.966+-.3040 | 2.154+-.0337 |

Table 1: Experimental results for learning with various percentage of missing values. The number in each cell indicates the mean and standard deviation of the KL divergence of the learned network to the true network from 5 different training sets (smaller is better). The variants of the BDe score are S, I, L, and N and they correspond to summation, integration, Laplace's, and linear approximations, respectively

In the reminder of this section, I will discuss how to apply these approximations for Dirichlet factors. Using (4), we have that:

$$\log F((N_{v_1}, \ldots, N_{v_l}))$$
$$= \log \Gamma(\sum_i N'_{v_i}) - \log \Gamma(\sum_i (N'_{v_i} + N(v_i)))$$
$$\quad + \sum_i (\log \Gamma(N'_{v_i} + N(v_i)) - \log \Gamma(N'_{v_i}))$$

It immediately follows, by linearity of expectations, that:

$$E[\log F((N_{v_1}, \ldots, N_{v_l}))]$$
$$= \sum_i E[\log \Gamma(N'_{v_i} + N(v_i))] -$$
$$\quad E[\log \Gamma(\sum_i (N'_{v_i} + N(v_i)))] + c,$$

where $c$ is some constant term that depends only on the prior.

As we can see, we can approximate each of the expectations individually. Since each one of these involves only one count, we will simplify notation somewhat. Assume that $\mu_i$ and $\sigma_i^2$ are the mean and variance of some count $N_i$. Also, let $N'_i$ be the prior count for the same event. Finally, let $m_i$, and $M_i$ be the minimal and maximal values that $N_i$ can take in the data. (These can be easily recorded during the computation of expected sufficient statistics.) We now consider three approximations to $E[\log \Gamma(N_i + N'_i)]$.

**Summation:** In this approximation, we iterate over the possible integral values of $N_i$ (from $m_i$ to $M_i$). For each value of $N_i$, we estimate the probability $p(N_i)$ using the Gaussian function, by integrating the range $[N_i - \frac{1}{2}, N_i + \frac{1}{2}]$ (for the extreme values $m_i$ and $M_i$, we also include also the volume of the tail of the the distribution). We then approximate $E[\log \Gamma(N_i + N'_i)]$ as

$$\sum_{N_i = m_i}^{M_i} \log \Gamma(N_i + N'_i)p(N_i).$$

This method does not scale when $N_i$ can take many values. However, I use is it a baseline to evaluate other approximations.

**Integration.** Using the continuous approximation to the sum above, we have that

$$E[\log \Gamma(N_i + N'_i)] \approx \int \log \Gamma^*(N_i + N'_i)\varphi(N_i : \mu_i, \sigma_i^2)dN_i,$$

where $\Gamma^*(\cdot)$ is the "truncated" $\Gamma(\cdot)$ function: $\Gamma^*(x) = \Gamma(x)$ if $x \in [m_i + N'_i, M_i + N'_i]$, $\Gamma^*(x) = \Gamma(m_i + N'_i)$ if $x < m_i + N'_i$, and $\Gamma^*(x) = \Gamma(M_i + N_i)$ if $x > M_i + N'_i$. This truncation is necessary since $\Gamma(x)$ grows to infinity as $x$ goes to 0. To evaluate this integral, we can use numerical integration procedures, called Hermite-Gaussian quadratures, that are particularly suitable for integrals of this form and can be encoded quite efficiently [1]. In the experiments described below, I use this integration procedure with 16 evaluation points. I suspect that it would suffice to use a smaller number of control points.

**Laplace's Approximation:** Here we approximate the integral of the Gaussian by finding the mode $m$ of the integrated function $\log \Gamma(x)\varphi(x : \mu_i + N'_i, \sigma_i^2)$. In my implementation, I find this value by binary search.

Using Laplace's approximation, we get that the integral is approximated by:

$$\log \Gamma(m)e^{-\frac{1}{2}\frac{(m-\mu_i-N'_i)^2}{\sigma_i^2}} \cdot$$
$$\left(1 - \sigma_i^2\left(\frac{(\log\Gamma)''(m)}{\log\Gamma(m)} - \left(\frac{(\log\Gamma)'(m)}{\log\Gamma(m)}\right)^2\right)\right)^{-\frac{1}{2}}$$

I use standard approximations (e.g., [1]) to compute the first and second derivatives of $\log \Gamma(\cdot)$.

# 5    EXPERIMENTAL RESULTS

## 5.1    METHODS

In this section, I describe results of experiments that indicate the effectiveness of the general approach and evaluate the alternative methods for computing scores discussed above.

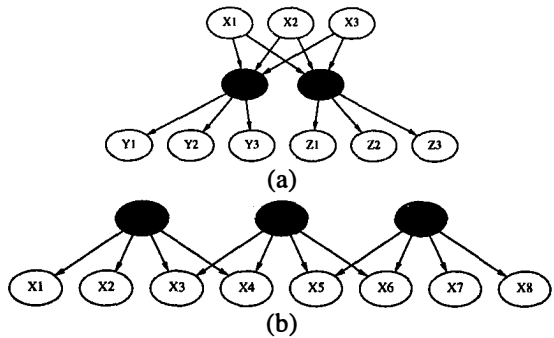

Figure 1: The networks used in learning with hidden variables. Shaded nodes correspond to hidden variables. (a) 3x1+1x3+3, (b) 3x8.

In addition, I also compare the resulting networks to networks learned using Structural EM with the BIC score (as described in [12]).

All the variants of this procedure use the same general architecture. There is a search module that performs greedy hill climbing search over network structures. To evaluate each network, this search procedure calls another module that is aware of the metric being used and of the current completion model. This module keeps a cache of expected sufficient statistics (and in the case of the Bayesian score, also variances and bounds) to avoid recomputations.

### 5.2  MISSING VALUES

Many real life data sets contain missing values. This poses a serious problem when learning models. When learning in presence of missing data, one has to be careful about the source of omissions. In general, omission of values can be informative. Thus, the learner should learn a model that maximize the probability of the *actual observations*, which includes the pattern of omissions. Learning procedures that attempt to score only the observable data, such as the one described here, ignore, in some sense, the missing values. This is justified when data is *missing at random* (MAR). I refer the interested reader to [23] for a detailed discussion of this issue. We can circumvent this requirement if we augment the data with indicator variables that record omissions, since the augmented data satisfies the MAR assumption. Thus, procedures, such as the one discussed here, are relevant also for dealing with data that is not missing at random.

In order to evaluate the Bayesian Structural EM procedure, I performed the following experiments that examine the degradation in performance of the learning procedures as a function of the percentage of missing values. In this experiment, I generated artificial training data from two networks: **alarm**—a network for intensive care patient monitoring [2] that has 37 variables, and **insurance**—a network for classifying car insurance applications [3] that has 26 variables. From each network I randomly sampled 5 training sets of different sizes, and then randomly removed values from each of these training sets to get training sets with varying percentage of missing values.

For each training set, the Bayesian and the BIC procedures were run from the same random initial networks with the same initial random seeds. These initial networks were random chain-like networks that connected all the variables. I evaluated the performance of the learned networks by measuring the KL divergence of the learned network to the generating network. The results are summarized in Table 1. As expected, there is a degradation in performance as the percent of missing values grows. We see that the Bayesian procedure consistently outperforms the BIC procedure, even though both use the same prior over parameters.

As we can see from these results, the summation approximation is consistently finding better networks. In some cases, it finds networks with as much as 60% small error than the linear approximation. This is especially noticeable for in the smaller training sets. The integration approximation performs slightly worst, but often significantly better than the linear approximation. These results match the hypothesis that the linear approximation is most unsuitable in small training sets. For larger training sets with small percent of missing values, we see that the linear approximation performs quite well, and often better than the Laplace approximation.

### 5.3  HIDDEN VARIABLES

In most domains, the observable variables describe only some of the relevant aspects of the world. This can have adverse effect on our learning procedure since the marginalization of hidden quantities can lead to a complex distribution over the observed variables. Thus, there is growing interest in learning networks that include one or more hidden variables. The Bayesian approach gives us the tools for learning a good structure with a fixed set of hidden variables. We still need an additional mechanism to choose how many hidden variables to add. This can be done using a simple loop, since we are now searching over a linear scale. The experiments in this section attempt to evaluate how good our procedure is in learning such hidden variables and how it compares with the BIC score which is easier to learn but over penalizes network structures.

In the experiments, I used two networks with binary variables: The first is **3x1+1x3+3** with the topology shown in Figure 1b. This network has hidden variables "meditating" between two groups of observed variables. The second is **3x8** with the topology shown in Figure 1b. Here all the variables seems to be correlated, although they are nicely separated by the hidden ones. I quantified these networks using parameters sampled from a Dirichlet distribution. For each sampled value for the parameters, I run a standard belief network learning procedure that used only the observable variables to see how "hard" it is to approximate the distribution. I then chose the parameter settings that led to the worst prediction on an independent test set.

I then sampled, from each network, training sets of sizes 500, 1000, 2000, and 4000 instances of the observable variables, and learned networks in the presence of 0, 1, 2, 3, or 4 hidden binary variables using the both the Bayesian Structural EM algorithm with the BDe metric with uniform prior, and the BIC Structural EM algorithm that used the same uniform prior over parameters. Both algorithms were started with the same set of initial network structure and randomized parameters.

In these experiments, the procedures are initialized by a structure in which all of the hidden variables are parents of each observable variable. (See [12] for motivation for the choice of this structure). As discussed above, both the Bayesian and the BIC versions of Structural EM can converge to local "structural" maxima. In the case of hidden variables, this phenomena is more pronounced than in the case of missing value. In these cases, the initial structure I use is often close to a local maxima in the search.

To escape from these local maxima, I use random perturbations. The procedure uses two forms of perturbations. In the first type of perturbations, a change the local neighborhood of the hidden variables is tried. This is done either by adding an edge to/from a hidden variable to another variable (which might be hidden), or reversing such an edge. After such a single edge change, the procedure restarts the Structural EM procedure with the new structure and runs until convergence. This is repeated where at each stage the procedure perturbs the best structure found so far. The procedure uses the Cheeseman-Stutz score [6, 7] to evaluate structures from different runs of Structural EM. (The BIC version uses the marginal BIC score.) This is repeated for up to five perturbations. After this type of perturbations is tried, the procedure applies the second type of perturbation, which is simply a random sequence of moves (edge addition, deletion and reversal). In the experiments the procedure applied 20 such changes. Then the procedure is restarted using the basic Structural EM procedure and the first type of perturbations. After 10 such random walks, or if the time limit is reached the procedure is terminated.

| # Hidden/ Method | 3x1+1x3+3 | | | | 3x8 | | | |
|---|---|---|---|---|---|---|---|---|
| | 500 | 1000 | 2000 | 4000 | 500 | 1000 | 2000 | 4000 |
| **0** | | | | | | | | |
| BDe | .1410+-.0246 | .0741+-.0205 | .0421+-.0123 | .0274+-.0046 | .1591+-.0226 | .0819+-.0104 | .0535+-.0057 | .0386+-.0046 |
| BIC | .1469+-.0274 | .0796+-.0233 | .0356+-.0035 | .0267+-.0029 | .1383+-.0192 | .0792+-.0108 | .0502+-.0035 | .0328+-.0038 |
| **1** | | | | | | | | |
| BDe (S) | .0964+-.0250 | .0384+-.0056 | .0240+-.0048 | .0159+-.0027 | .1063+-.0182 | .0423+-.0138 | .0419+-.0028 | .0261+-.0011 |
| BDe (I) | .0698+-.0195 | .0431+-.0107 | .0222+-.0023 | .0165+-.0011 | .1085+-.0241 | .0438+-.0111 | .0319+-.0060 | .0235+-.0043 |
| BDe (La) | .0831+-.0132 | .0374+-.0041 | .0214+-.0027 | .0151+-.0022 | .0892+-.0235 | .0513+-.0122 | .0348+-.0099 | .0224+-.0058 |
| BDe (Li) | .0920+-.0201 | .0409+-.0088 | .0241+-.0058 | .0144+-.0026 | .1078+-.0138 | .0443+-.0093 | .0358+-.0056 | .0227+-.0060 |
| BIC | .0929+-.0101 | .0590+-.0166 | .0224+-.0028 | .0182+-.0024 | .1152+-.0213 | .0635+-.0092 | .0294+-.0051 | .0247+-.0076 |
| **2** | | | | | | | | |
| BDe (S) | .0720+-.0249 | .0304+-.0037 | .0174+-.0039 | .0100+-.0034 | .0785+-.0223 | .0422+-.0112 | .0209+-.0024 | .0163+-.0053 |
| BDe (I) | .0731+-.0321 | .0323+-.0051 | .0147+-.0046 | .0098+-.0022 | .0907+-.0244 | .0364+-.0085 | .0228+-.0031 | .0134+-.0057 |
| BDe (La) | .0702+-.0307 | .0403+-.0088 | .0127+-.0039 | .0113+-.0037 | .0769+-.0336 | .0485+-.0212 | .0221+-.0038 | .0157+-.0030 |
| BDe (Li) | .0646+-.0175 | .0290+-.0043 | .0134+-.0042 | .0070+-.0020 | .0619+-.0209 | .0344+-.0054 | .0196+-.0021 | .0165+-.0017 |
| BIC | .0952+-.0259 | .0333+-.0035 | .0133+-.0028 | .0082+-.0019 | .1074+-.0494 | .0428+-.0069 | .0209+-.0015 | .0204+-.0035 |
| **3** | | | | | | | | |
| BDe (S) | .0875+-.0282 | .0504+-.0221 | .0253+-.0075 | .0158+-.0021 | .0386+-.0176 | .0365+-.0168 | .0248+-.0095 | .0158+-.0042 |
| BDe (I) | .0889+-.0245 | .0382+-.0062 | .0229+-.0099 | .0100+-.0044 | .0516+-.0165 | .0409+-.0251 | .0193+-.0099 | .0106+-.0040 |
| BDe (La) | .1079+-.0157 | .0335+-.0153 | .0166+-.0066 | .0138+-.0050 | .0465+-.0156 | .0274+-.0094 | .0148+-.0084 | .0123+-.0068 |
| BDe (Li) | .1058+-.0215 | .0298+-.0080 | .0198+-.0031 | .0143+-.0052 | .0481+-.0268 | .0276+-.0053 | .0184+-.0073 | .0136+-.0056 |
| BIC | .1108+-.0383 | .0574+-.0203 | .0143+-.0044 | .0096+-.0040 | .0679+-.0217 | .0185+-.0073 | .0082+-.0020 | .0073+-.0048 |
| **4** | | | | | | | | |
| BDe (S) | .0678+-.0179 | .0676+-.0157 | .0615+-.0167 | .0263+-.0089 | .0628+-.0147 | .0673+-.0063 | .0309+-.0042 | .0154+-.0032 |
| BDe (I) | .0942+-.0217 | .0847+-.0296 | .0365+-.0196 | .0206+-.0065 | .0564+-.0260 | .0448+-.0160 | .0321+-.0096 | .0145+-.0040 |
| BDe (La) | .0880+-.0163 | .0357+-.0159 | .0365+-.0098 | .0220+-.0053 | .0458+-.0189 | .0372+-.0096 | .0262+-.0065 | .0158+-.0027 |
| BDe (Li) | .1105+-.0308 | .0373+-.0108 | .0228+-.0047 | .0125+-.0016 | .0594+-.0230 | .0266+-.0088 | .0185+-.0075 | .0133+-.0045 |
| BIC | .1181+-.0131 | .0628+-.0186 | .0260+-.0087 | .0162+-.0105 | .0715+-.0252 | .0279+-.0128 | .0151+-.0057 | .0082+-.0033 |

Table 2: Performance on an independent test set for the networks learned with hidden variables using the BDe and BIC scores. The reported numbers correspond to the difference in log loss on the test set between the generating distribution and learned distributions. The mean and standard deviation of this quantity for run on 5 data sets are reported. The labels of the rows indicate the number of hidden variables that were learned and the procedure used.

The results summarized in Table 2, show that the variants of the Bayesian procedure usually make better predictions than the BIC score, but not always. Also, the performance of the linear approximation is often better than other approximations. The main explanation for both of these discrepancies from the missing data case, is that in these learning problems the main improvements where achieved by runs that where initialized by the "right" random perturbations. Since, all the runs were terminated after 30 CPU minutes, the runs with the BIC score and the BDe with linear approximation have gone through many more random restarts than the other runs. This is most noticeable in the cases where there are more hidden variables, since they require many score evaluations for factors with incomplete data and the search space they define contain more local maxima. The structures learned where also quite close to the original structure. Due to space restrictions, I cannot elaborate on this here.

## 6   DISCUSSION

In this paper, I described a new approach for Bayesian model selection in belief networks and related models. I believe that this approach is exciting since it attempts to directly optimize the true Bayesian score within EM iterations. The paper describes a framework for building algorithms that learn from incomplete data. This framework provides some guarantees, but leaves open such issues as the collection of sufficient statistics and the computation of the expected score for each factor. These details can be filled in for each class of models.

There is quite a bit of related work on learning from incomplete data. The general idea of interleaving structure search with EM-like iteration appeared in several papers. The first Structural EM paper, Friedman [12] introduced the framework and established the first formal convergence results. Singh [25] had a similar insight although his procedure is somewhat different. Like the Structural EM procedure, his procedure is iterative. In each iteration, it generates $k$ joint assignments to all missing values using the best model from previous iterations. His procedure then invokes the learning procedure of Cooper and Herskovits [9] on each one of the completed datasets. Finally, Singh's procedure merges the learned networks, trains parameters for this merged network using standard EM procedure, and reiterates. This approach can be interpreted as a stochastic approximation of Structural EM. The analysis of this paper gives insight into the limiting behavior of Singh's algorithm. More precisely, by using $k$ completed datasets, Singh approximates the expectation of the score. However, instead of combining these estimates within a single search procedure, Singh searches for structures independently on each one of the completed datasets. This leads to various complications, such as the need to merge the learned networks.

Some variants of Structural EM have been proposed by Meila and Jordan [21] and Thiesson et al. [27]. Both of these variants learn multinets in which the selector variable is hidden (these can be thought of mixtures of Bayesian networks). Meila and Jordan learn multinets in which each network is a Chow tree. They exploit this restriction to collect all required statistics in one pass at each iteration. Although they do not provide any formal treatment of their procedure, the analysis of [12] directly applies to their approach, and shows that their procedure will converge to a local maximum. Thiesson et al. [27] aim to learn general multinets using the Cheeseman-Stutz score [6]. By examining approximations to this score they motivate a learning algorithm that, in the terminology of this paper, can be seen as an instance of Factored-Bayesian-SEM, using the linear approximation, applied to multinets. Thiesson et al. use an efficient method for caching expected statistics when most of the variables of interest are Gaussian, that can answer all queries during the structure search after a single pass on the training data at each iteration. The analysis in this paper directly applies to their approach.

One restriction of the Structural EM algorithm is that it focuses on learning a single model. In practice, we often want to use a *committee* of several high scoring models for prediction. Such committees can provide a better approximation of Eq. (1) and ensure that we do not commit to the particulars of a single model when the evidence also supports other models. Both Meila and Jordan, and Thiesson et al. attempt to approximate such committees by learning *mixture models*, where each mixture component is a Bayesian network. Nonetheless, they are learning a MAP model, in a larger class of models. This might be useful, if the source of the data can be better described by a mixture. However, it does not address the dependency on a single model.

Alternatively, we might attempt to directly follow the basic Bayesian principle as formulated in Eq. (1), and perform Bayesian model averaging. In this approach, members of the committee are weighted by their posterior probability. It turns out that we can use a variant of Bayesian Structural EM to learn Bayesian committees. Roughly speaking, we can run Bayesian Structural EM where the "current" candidate at each stage is a Bayesian committee of models (i.e., each model is weighted by its posterior probability). Then, at each iteration we choose the $k$ models that have the highest expected score given the current committee. The formal treatment of this idea is somewhat more complex, and is the topic of current research.

There are several other issues that require additional understanding. In particular, although I provided convergence proofs for the abstract version of the algorithm, it is still not clear whether these proofs apply given the approximations need to perform this algorithm in practice. Empirical experience shows that the procedure does consistently converge. However, better theoretical understanding is called for.

An additional aspect glossed over in this presentation is the computation of the expected statistics. This requires large number of computations during learning. This is the main bottleneck in applying this technique to large scale domains. It is clear that we should be able to improve the standard inference procedures by exploiting the fact that we are evaluating the same set of queries over large number of instances. Moreover, stochastic simulation seems an attractive approach to examine in this context, since we can use the same sample to evaluate many queries. This, however, requires a more careful analysis of the effect of the noise in the estimation on the convergence properties of the algorithm. Finally, it would be interesting to understand if it is possible to combine variational approaches (e.g., [24]) with this type of learning procedures.

Another major open question is how to decide, in an in-telligent fashion, on the number of hidden variables. Right now, the approach used in this paper (and in [12, 21, 27]) is to learn models with 1 hidden variable, 2 hidden variables, etc., and then to select the network with the highest score. This is clearly a blind approach. Moreover, the qualitative model learned with a hidden variable depends on the initial structure used by the Structural EM procedure. Current research examines how to combine the Structural EM procedure with constraint-based approaches, such as these of [26] that learn constraints as to the possible positions of hidden variables, to guide the introduction of hidden variables during the search.

## Acknowledgments

I am grateful to Danny Geiger, Moises Goldszmidt, Daphne Koller, Kevin Murphy, Ron Parr, Stuart Russell, and Zohar Yakhini for useful discussions relating to this work. I would like to thank an anonymous referee, whose comments prompted lead me to investigate the appropriateness of the linear approximation in more detail. Some of this work was done while I was at SRI International. This research was supported by ARO under grant number DAAH04-96-1-0341 and by ONR under grant number N00014-97-1-0941.